\documentclass[letterpaper, 10 pt, conference]{ieeeconf}  
\usepackage[utf8]{inputenc}
\usepackage[parfill]{parskip}
\usepackage{amsfonts}
\usepackage{graphicx}
\usepackage{booktabs}
\usepackage{amsmath}
\usepackage[final]{pdfpages}

\IEEEoverridecommandlockouts                              

\title{\LARGE \bf
AptSim2Real: Approximately-Paired Sim-to-Real Image Translation }

\author{{\textbf {Charles Y Zhang}$^*$} \\
{ \small University of Waterloo, Canada}\\
 {\tt\small cy9zhang@uwaterloo.ca}
\and
{\textbf {Ashish Shrivastava}} \\
{ \small Cruise LLC, United States}\\
{\tt\small ashish.shrivastava@getcruise.com}
\thanks{$^*$Work done during internship at Cruise LLC.}
}

\graphicspath{{figures/}} 

\begin{document}
   
\maketitle
\thispagestyle{empty}
\pagestyle{empty}

\begin{abstract}



Advancements in graphics technology has increased the use of simulated data for training machine learning models. 
However, the simulated data often differs from real-world data, creating a distribution gap that can decrease the efficacy of models trained on simulation data in real-world applications.
To mitigate this gap, sim-to-real domain transfer modifies simulated images to better match real-world data, enabling the effective use of simulation data in model training.

Sim-to-real transfer utilizes image translation methods, which are divided into two main categories: paired and unpaired image-to-image translation. Paired image translation requires a perfect pixel match, making it difficult to apply in practice due to the lack of pixel-wise correspondence between simulation and real-world data.
Unpaired image translation, while more suitable for sim-to-real transfer, is still challenging to learn for complex natural scenes. 
To address these challenges, we propose a third category: approximately-paired sim-to-real translation, where the source and target images do not need to be exactly paired.
Our approximately-paired method, AptSim2Real, exploits the fact that simulators can generate scenes loosely resembling real-world scenes in terms of lighting, environment, and composition. Our novel training strategy results in significant qualitative and quantitative improvements, with up to a $24\%$ improvement in FID score compared to the state-of-the-art unpaired image-translation methods.

\end{abstract}

\section{Introduction}

Recent improvements in simulation technology have made it a vital component in the training and validation of machine learning models for robotics applications, especially for tasks where real world data is costly or impossible to collect.
For example, OpenAI demonstrated that models trained only in simulation can be used to solve a Rubik's cube with a robotic hand \cite{akkaya2019solving}, researchers at ETH Zurich were able to learn complex quadrupedal locomotion using simulation data \cite{lee2020learning}, and Microsoft recently released a simulation dataset for learning localization models for drones \cite{wang2020tartanair}.
In particular, training visual detection and understanding algorithms on synthetic image data can produce immense gains for a critical part of a robotic system.
Previous studies have shown that enhancing the quality of synthetic data using sim-to-real domain adaptation techniques improves the performance of downstream models on real data. (\cite{wood2021fake, haiderbhai2022dvrksim2real, simgan2017}). 

Current research in sim-to-real domain adaptation focuses on two main approaches: paired image translation and unpaired image translation. Paired methods train a model to translate images between source and target domains where each source image has a corresponding paired target image with pixel-wise correspondences. 
This approach results in high accuracy, but paired data is often difficult and expensive to obtain. 
In contrast, the unpaired image translation methods train a model with no correspondences between source and target images.
This approach is more challenging but benefits from a higher availability of data. 


\begin{figure}[!t]
    \centering
    \includegraphics[width=0.95\columnwidth]{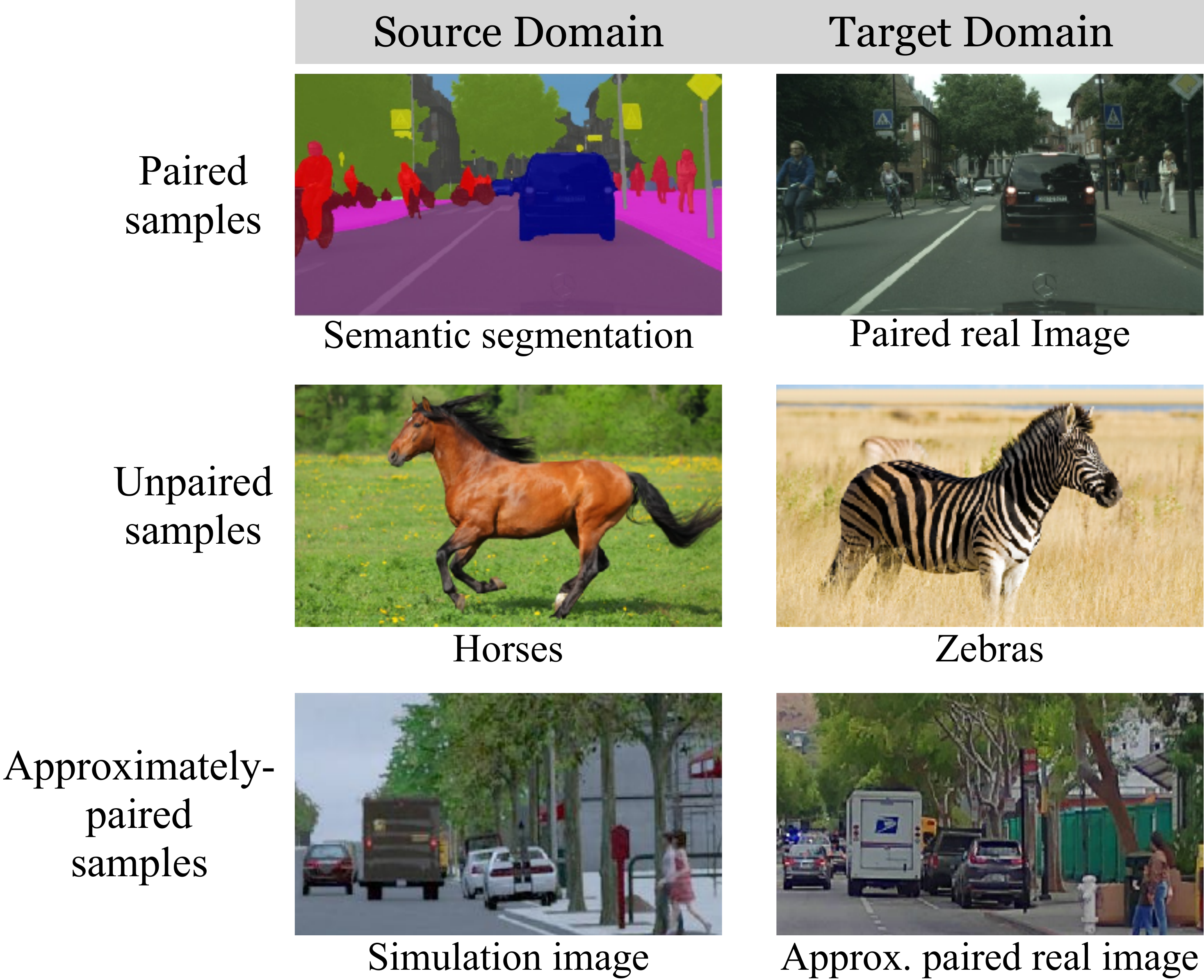}
    \caption{
    \textbf{AptSim2Real} differs from traditional paired or unpaired methods by rendering simulated images with similar camera pose, background, lighting, and scene composition to real images, resulting in an approximate pairing between the two.
The approximately-paired samples enable the use of stronger supervision compared to unpaired methods during the learning process, while still retaining data scalability, a challenge faced by traditional paired methods.
\vspace{-0.2in}
    }
    \label{fig:apt}
\end{figure}

In this paper, we introduce a novel approach to sim-to-real domain transfer called \emph{a}pproximately-\emph{p}aired image \emph{t}ranslation  (AptSim2Real).
Unlike paired or unpaired methods, this approach utilizes ``approximately-paired" data that shares contextual information such as camera pose, map location, scene composition, and lighting while allowing some variations in assets, textures, appearance, and shapes (see Figure~\ref{fig:apt}). 
For each real image, we use metadata and label information to generate a corresponding simulated image in a graphics engine.
Simulated images mirror the camera pose of real images and are generated using assets and models similar to those in the real image, a procedurally generated background, and matching lighting.
This approach allows for scalable data generation, like unpaired translation, while additionally providing some pairing between the source and target images that the model can take advantage of. 
We demonstrate that leveraging approximate pairings between source and target domain samples can lead to a superior model architecture and better performance compared to unpaired translation methods.



The goal of sim-to-real training is to mimic the real images as closely as possible. 
Due to limitations of simulation technology, there is a significant distribution gap between the simulated and real images.
Our method bridges this gap by exploiting approximately-paired data using style mixing \cite{pmlr-style-equalization}.
During the training process, a style encoder network learns to encode the ``realism gap" between a simulated image and its corresponding approximately-paired real image into a style difference feature. 
This style difference is used as an additional input to a generator network to guide image translation and improve the realism of the simulated image.
While prior image translation methods such as \cite{simgan2017, park2020cut, CycleGAN2017} encode the style difference within their parameters, we take a different approach by presenting this difference as an additional style input. 
This alternative strategy not only simplifies the translation task but also provides greater flexibility and control over the style transfer process.
Note that approximate pairing between the simulated input image and the generated image is crucial for computing the style difference and, therefore, is central to our architecture design.
Following are our main contributions:

\begin{itemize}
    \item We propose approximately-paired image translation -- a novel image translation category for improving the realism of simulated images.

   \item We propose a novel training strategy that leverages the approximate-pairing between the source and target domain by utilizing the latest development in GAN methods.

    \item We conduct extensive qualitative and quantitative experiments to demonstrate that approximately paired images translation can be used to produce results significantly more realistic than existing unpaired methods.


\end{itemize}



\section{Related Works}

Generative Adversarial Networks (GANs)~\cite{gan_2014} employ a critic, also known as a discriminator, to differentiate between real and fake data inputs.
GANs are widely used in image translation, as training a generative model to fool the critic (discriminator) has proven effective in closing distribution gaps.
As GAN methods have advanced (~\cite{biggan_iclr_2019, adain_cvpr_2017, msggan_cvpr_2020, progressive_gan_2017, stylegan, WasserstineGAN_2017, gan_2014, SAGAN_2018, stackgan_2018, stackgan_pp_2017, cogan_2016, lsgan_2017, karras2021_stylegan3aliasfree}), the performance of image translation has also improved.

Paired image translation uses an input, typically in the form of a label such as a semantic segmentation map or edge map, to generate a new output image~\cite{isola2017_pix2pix, Qu_2019_CVPR, cond_gan_cvpr_2020, instaformer_2022, Scribbler_cvpr_2017, attr_layout_to_img_2016}. 
Paired image translation uses pixel-wise correspondences to jointly optimize an L1 loss between the predicted and target images alongside an adversarial loss to achieve high accuracy.

Unpaired image translation lacks the pixel-wise correspondences of paired image translation and instead utilizes alternative loss functions. For example, SimGAN~\cite{simgan2017}  regularizes the difference between the input and output of the model, CycleGAN~\cite{CycleGAN2017} optimizes for cycle-consistency, and CUT~\cite{park2020cut} maximizes mutual information between the input and output of the generator. 
While other methods have been proposed(~\cite{CycleGAN2017, park2020cut, simgan2017, unsup_img2img_gen_prior_2022, unsup_im2im_dense_consistensy_2022}), CycleGAN and CUT remain popular choices for unpaired image translation.

Combining paired and unpaired image translation has also been proposed for datasets that have a mix of paired and unpaired images~\cite{paired_and_unpaired_accv_2018,Mustafa_eccv_2020}. 
In contrast, our method does not rely on pixel-wise correspondences, generates approximately-paired synthetic images through simulation, and leverages these approximate pairings with state-of-the-art GAN techniques.
\section{Method}

\begin{figure*}[htp]
    \centering
    \includegraphics[width=\textwidth]{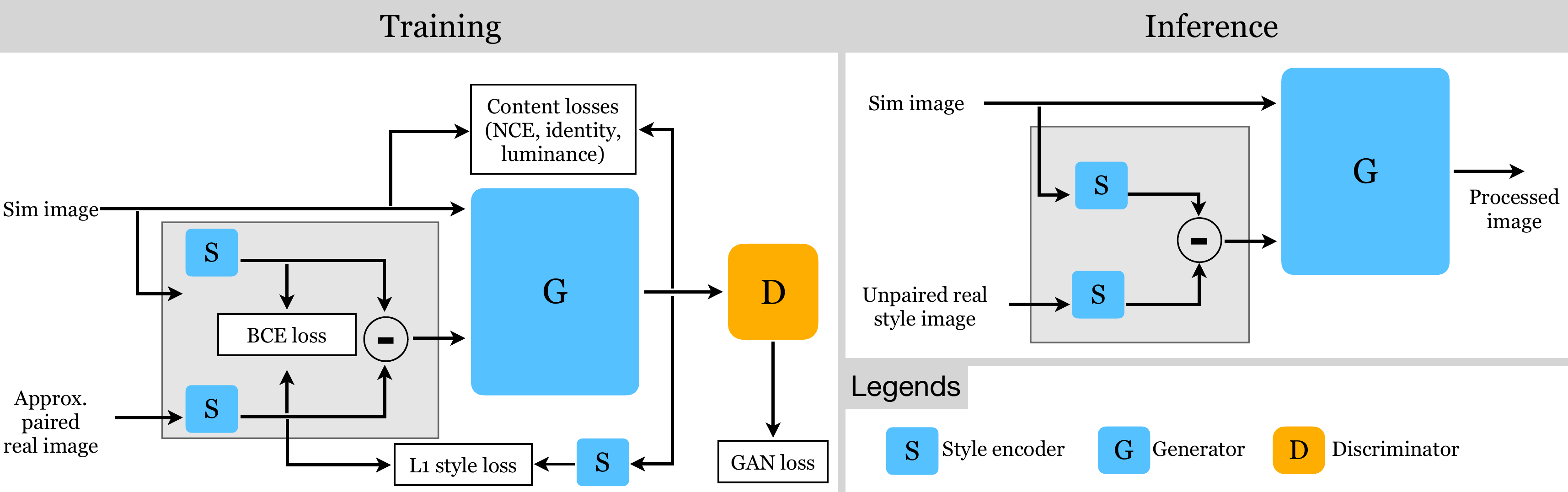}
    \caption{\textbf{Architecture Overview.} 
    To leverage approximately-paired data, AptSim2Real trains a style encoder, ${S}$, to extract a style code from the approximately paired inputs. During training, we condition the generator, ${G}$, on the difference in input and target style. Given that the simulation and real images are almost identical in content (as they are approximately-paired), the difference in style features reflects the realism gap between the images as any content feature information is cancelled. Content regularization losses and pooling layers in the style encoder are employed to additionally minimize content leakage. During inference, a real image with a similar style as the sim input is selected from the training data as the target style reference. 
    }
    \label{fig:architecture}
\end{figure*}


\subsection{Generating Approximately-paired Images}  

To learn a mapping $G: \boldsymbol x \rightarrow \boldsymbol y$ from simulated images, $\boldsymbol x$, to real images, $\boldsymbol y$, 
  we rely on approximately-paired image pairs, $(\boldsymbol x, \boldsymbol y)$. 
Unlike paired image translation~\cite{isola2017_pix2pix}, our pairs do not require pixel-wise alignment.
Instead, we sample a real image $\boldsymbol y_i$ from the real dataset and generate a corresponding simulated image by utilizing the metadata and the label information present in the image.
Since our real images are collected from sensors on an autonomous vehicle, we have a lot of metadata information including pose of the sensor, time of day, location, lighting, and weather.
Using this contextual information, we render a simulated image $\boldsymbol x_i$ in a graphics engine. 
We create a simulated version of each real object by selecting the corresponding asset from an asset library. 
Our selection process takes into account the label and contextual information present in the real image.
Given the pose information of the autonomous vehicle, we select the background of the simulation scene from a procedurally generated 3D map. 
The lighting and weather conditions are matched by selecting an environment map that reflects these properties.
Finally, we construct the completed 3D scene in a graphics engine by combining the selected assets, background, and lighting. This produces a simulated image that mirrors the camera pose of the real image.
We repeat this process for each real image and generate a dataset of $N$ approximately-paired images $\{\boldsymbol x_i, \boldsymbol y_i\}_{i=1}^N$.

This scalable image generation process yields a dataset consisting of image pairs that share attributes such as object size, type, background, scene composition, lighting, and camera pose.
Much like unpaired data, Approximately-paired data is cheap and efficient to generate with minimal manual labeling and curation needed.
In contrast, paired data requires substantial manual curation due to the need for pixel-level correspondence.
However, like paired data, approximately-paired data incorporates contextual information that can guide the model and simplify the learning process, resulting in faster training and improved accuracy.

During training, the model is fed with approximately paired images, ${\boldsymbol{x}, \boldsymbol{y}}$. 
This pairing makes data generation more straightforward and improves supervision. 
At inference, a style image $\boldsymbol{y}$ possessing the desired style is selected from the training dataset.




\subsection{Model details}
SimApt2Real is formulated as a one-sided generative adversarial network~\cite{onesided_gan_2017}
and consists of a generator $G$, a style encoder $S$, and a discriminator $D$. 
The goal of the network is to learn a generator $G$ which can take as input simulated images $\boldsymbol x$ and generate realistic images reflecting the style of the approximately-paired style image. 
The architecture leverages approximately paired data to maximize its performance, as shown in the overview in Figure \ref{fig:architecture}.

\textbf{Style encoder}. The style encoder $S$ takes an image $\boldsymbol {x}$ as an input and generates a style code $\boldsymbol s = S(\boldsymbol x)$. 
Given the approximately-paired style input $\boldsymbol y$, we compute the style difference, $\triangle{\boldsymbol{s}} = S(\boldsymbol y) - S(\boldsymbol x)$. The style difference is used as an additional input to the generator that identifies what realism differences that must be corrected:
$$\boldsymbol {\hat y} = G(\boldsymbol x, \triangle \boldsymbol{s}).$$
Approximate pairing is critical for the style encoder design as we assume that the simulation and real images are almost identical in content, meaning that the difference in style features will reflect solely the realism gap. Any content-level feature differences will be cancelled when the style codes are subtracted.

To prevent the generator from over-fitting to the style input during training, we remove spatial content information from the style inpuet $\boldsymbol y$ by incorporating pooling layers in the style network.
We use multiple content losses between the input simulated image and the generated image to ensure that the generated image accurately reflects the content of the input image. 
These steps help to ensure that the generator is not solely focused on replicating the style of the real images but is also generating content accurate to the input image.


\textbf{Generator}. The generator $G$ consists of multiple ResNet blocks~\cite{Resnet_He2015} and translates an input simulated image $\boldsymbol{x}$ into a realistic output image $\boldsymbol{\hat y} = G(\boldsymbol{x}, \triangle \boldsymbol{s})$ guided by the style difference, $\triangle \boldsymbol{s}$, which is a $d$-dimensional learned latent code that represents the ``realism gap" between the simulated image and the corresponding approximately-paired style input.
We ``modulate" and ``demodulate" the weights of the convolution layers in the generator, $G$, using this style difference vector, $\triangle \boldsymbol{s}$, similar to the method proposed in~\cite{Karras2019stylegan2}.
For each convolution layer, a linear layer is used to map the vector $\triangle \boldsymbol{s}$ so that the new dimension is same as the number of input feature maps in the convolution layer. 
Let $s_i$ be the $i^\text{th}$ element of the mapped style vector.
The convolution weights corresponding to the $i^\text{th}$ input feature map are ``modulated" by multiplying it with $s_i$.
Each modulated convolution weight is then ``demodulated" to normalize the output feature map.
These modified convolutions are used to replace a typical instance norm in the generator \cite{Karras2019stylegan2} and allow us to control the weights of a convolution given the style difference $\triangle \boldsymbol{s}$. 
We refer readers to equations 1, 2, and 3 in \cite{Karras2019stylegan2} for details.

\textbf{Discriminator}. The discriminator, $D$, is trained to distinguish between real images, $\boldsymbol{y}$, and generated images, $\boldsymbol{\hat y}$, in a manner similar to standard GAN approaches~\cite{lsgan_2017}. 
Following the approach in \cite{simgan2017}, the discriminator is modeled with multiple convolution layers, producing a feature map of a lower resolution than the input image. 
Each location in the feature map corresponds to a patch in the input image, and the size of each patch is determined by the size of the receptive field of the output features. 
By classifying each location of the output feature map as real or fake, the discriminator classifies each patch in the input.

\subsection{Training Losses}

Following standard GAN methods, we alternate between solving the following two optimization problems:
\begin{align*}
&\min_{D} \; \mathbb{E}_{\boldsymbol x,\boldsymbol y}\left[(D(\boldsymbol y) - 1)^2 + (D(G(\boldsymbol x, \triangle \boldsymbol{s}))-0)^2\right], \nonumber \\ 
&\min_{S, G} \mathcal{L}_{total}(S, G).
\end{align*}
Here, we employ the least squares adversarial loss~\cite{lsgan_2017}, where the discriminator is trained to predict $0$ for generated images and $1$ for real images. 
The total loss for optimizing the generator ($G$) and style encoder ($S$), $\mathcal{L}_{total}$, comprises of an adversarial loss aimed at fooling the discriminator. 
Next, we specify the various components that make up the total loss function.

\textbf{Adversarial loss}. The adversarial loss aims to deceive the discriminator into classifying the generated images as real, meaning that the discriminator should predict $1$ for the generated images.
Given $\triangle \boldsymbol{s} = S(\boldsymbol y) - S(\boldsymbol x)$, 
\begin{align*}
\mathcal{L}_{GAN} = \mathbb{E}_{\boldsymbol x,\boldsymbol y}\left[ (D(G(\boldsymbol x, \triangle \boldsymbol{ s})) - 1)^2\right].  
\end{align*}

\vspace{-0.12in}
\textbf{Style Reconstruction loss.} To ensure style similarity between the generated images, $\hat {\boldsymbol y}$, and target images, $\boldsymbol y$, we minimize the $\ell_1$ loss between the style representations of $\hat {\boldsymbol y}$ and $\boldsymbol y$ as follows:
\begin{align*}
\mathcal{L}_{sty} = \mathbb{E}_{\boldsymbol y}\left[\|S(\boldsymbol y) - S(\hat{\boldsymbol y})\|_1\right].
\end{align*}

Note that the style encoder can learn to output a constant vector to minimize this loss.
To circumvent this issue, we utilize the BCE loss (explained below) to distinguish between real image styles and simulated image styles.

\textbf{Style classification (BCE) loss.} The style encoder's key property is to differentiate the styles of approximately paired simulated and real images. 
To enforce this, we use a binary cross-entropy (BCE) loss to separate the style vectors of simulated and real images.
We train a linear classifier that maps the style vector to a scalar, producing $0$ for simulated style vectors and $1$ for real style vectors. 
The linear classifier is represented by a $d \times 1$ matrix $W_s$. 
The style classification loss is defined as follows:
\begin{align*}
    \mathcal{L}_{bce} = \mathbb{E}_{\boldsymbol x,\boldsymbol y}\left[(\log (f(W_s(S(\boldsymbol y))) + \log(1-f(W_s(S(\boldsymbol x))\right],
\end{align*}

\vspace{-0.11in}
where $f(.)$ is a Sigmoid function: $f(x) = 1/(1+e^{-x})$.
For improved stability and convergence during training, $W_s$ is pre-trained for a single epoch and then fine-tuned with a lower learning rate in conjunction with the other losses during the overall training process.
We found that the BCE loss and the style reconstruction loss are sufficient to capture the style differences between simulation and real images and that the style vectors are not ignored by the generator model.

\textbf{Content NCE loss}. We leverage PatchNCE loss~\cite{park2020cut} to enforce content similarity between $\boldsymbol x$ and $\hat {\boldsymbol y}$. 
Our generator consists of an encoder and a decoder component, i.e. $G(\boldsymbol x, \triangle \boldsymbol s) = G_{dec}(G_{enc}(\boldsymbol x, \triangle \boldsymbol s))$.
Using $G_{enc}$, we compute the encoded feature maps for the input image, $\boldsymbol x_{enc}$, and the generated image, $\hat {\boldsymbol y}_{enc}$, as follows: $\boldsymbol{x}_{enc} = G_{enc}(\boldsymbol{x}, S(\boldsymbol y) - S(\boldsymbol x))$ and $\hat{\boldsymbol{y}}_{enc} = G_{enc}(\hat{\boldsymbol{y}}, S(\boldsymbol y) - S(\hat{\boldsymbol y}))$.
Next, we sample the encoded feature vectors $\hat {\boldsymbol v}$ from the feature map $\hat {\boldsymbol y}_{enc}$ and the corresponding 
positive and negative feature vectors, $\boldsymbol v^+$ and $\boldsymbol v^-$, from the simulated feature map $\boldsymbol{x}_{enc}$. 
Note that $\boldsymbol v^+$ and $\hat {\boldsymbol v}$ correspond to the same spatial location while  $\boldsymbol v^-$ and  $\hat {\boldsymbol v}$ correspond to different spatial locations in the feature map.

To be precise, we first randomly sample $n$ indices from the generated feature map, $\hat {\boldsymbol y}$.
Let this set of indices be represented as $K = \{k_1, \dots, k_n\}$.
For each spatial location $k \in K $, we sample $\hat {\boldsymbol v}_{k} = \hat {\boldsymbol y}_{enc}(k)$, positive feature ${\boldsymbol v}^+_k = \hat {\boldsymbol x}_{enc}(k)$, negative feature ${\boldsymbol v}^-_k = \hat {\boldsymbol x}_{enc}(m)$ where $m \in K$ and $m \neq k$. 
In all our experiments, we set $n=256$.
Using the triplet $(\hat {\boldsymbol v_k}, {\boldsymbol v_k}^+, {\boldsymbol v_k}^-)$, the NCE loss is defined as follows:
\begin{align*}
& \ell_{nce}^{(k)} = - \log \left[ \frac{\exp(\hat {\boldsymbol v_k} \cdot {\boldsymbol v_k}^+ / \tau)} {\exp (\hat {\boldsymbol v_k} \cdot {\boldsymbol v_k}^+ / \tau + \sum_{n=1}^{N} \exp(\hat {\boldsymbol v_k} \cdot {\boldsymbol v_k}^-/\tau))}\right],   \\
& \mathcal{L}_{NCE} = \mathbb{E}_{\boldsymbol x, \boldsymbol y}\left[ \sum_{k \in K} \ell_{nce}^{(k)} \right],
\end{align*}
where the sum is over all the features sampled from an image (i.e. $(\hat {\boldsymbol v}, {\boldsymbol v}^+, {\boldsymbol v}^-)$ triplets).


\textbf{Content identity loss}. Since we modify the input image $\boldsymbol x$ using the style difference, a zero style difference should result in no modification to the input.
To ensure this identity, we add an $\ell_1$ loss between the input image and the generated image with a zero style difference, expressed as follows:
$$\mathcal{L}_{idt} = \mathbb{E}_{\boldsymbol x}\left[\left\|G(\boldsymbol x, \boldsymbol 0) - \boldsymbol x\right\|_1\right].$$




\vspace{-0.1in}
\textbf{Content luminance loss}. 
This loss function aims to preserve the content of the image, such as object locations, shapes, and edges.
A naive approach to preserve the content would be to make sure that the luminance component of the simulated and generated images is identical. 
However, this constraint would prevent the model from making changes to style properties that are dependent on luminance, such as noise, overall brightness, and contrast. 
To overcome this limitation, the difference between normalized luminance patches of size 16x16 is minimized, as described below:
\begin{align*}
    \mathcal{L}_{lum} = \mathbb{E}_{\boldsymbol x, \boldsymbol y}\Bigg[\left\|\frac{\bar L_{\boldsymbol x} - \mu(\bar L_{\boldsymbol x})}{\sigma(\bar L_{\boldsymbol x})} - \frac{\bar L_{\hat{\boldsymbol y}} - \mu(\bar L_{\hat{\boldsymbol y}})}{\sigma(\bar L_{\hat{\boldsymbol y}})}\right\|_1\Bigg],
\end{align*}

where the luminance of an RGB image, $\boldsymbol x$, is calculated as a weighted combination of its red, green, and blue channels, with $L_{\boldsymbol x} = 0.299 R_{\boldsymbol x} + 0.587 G_{\boldsymbol x} + 0.114 B_{\boldsymbol x}$. 
$\bar L_{\boldsymbol x} $  (or $\bar L_{\boldsymbol y} $ ) is output of the $16 \times 16$ average pooling operation with a stride of $16$ on the input luminance $L_{\boldsymbol x}$ (or the generated luminance $L_{\boldsymbol y}$).
The $\mu(\cdot)$ and $\sigma(\cdot)$ denote the mean and the standard deviation functions, respectively.

\textbf{Total loss}. We sum all the above losses for $S$ and $G$ with the following combination weights,
\begin{align}
\mathcal{L}_{total}(S, G) = & \mathcal{L}_{GAN} + \lambda^{cls} \mathcal{L}_{bce}+ \lambda^{sty} \mathcal{L}_{sty} \nonumber \\
& + \lambda^{nce} \mathcal{L}_{nce} + \lambda^{idt} \mathcal{L}_{idt} + \lambda^{lum} \mathcal{L}_{lum}. \nonumber
\end{align}
In all our experiments, we set $\lambda^{cls}=1.0, \lambda^{sty} = 1.0, \lambda^{nce} = 0.5, \lambda^{idt} = 0.5, \lambda^{lum} = 0.1$.

\textbf{Note on approximate-pairing}. It is worth noting that the computation of style difference, $\triangle \boldsymbol s $, assumes that majority of the content in the simulated input is closely matched with the content in the real style input. 
If the input is not approximately-paired, then the computed $\triangle \boldsymbol s $ would not only capture the style difference but also the content difference. 
Moreover, using approximately-paired data allows for generated and approximately-paired real samples to be included in the same mini-batch for the discriminator training, which enables the adversarial loss to better focus on style differences and encourage the generator to produce outputs that more closely match the target style.

\section{Experiments}

\begin{figure*}[htp]
    \centering
    \includegraphics[width=\textwidth]{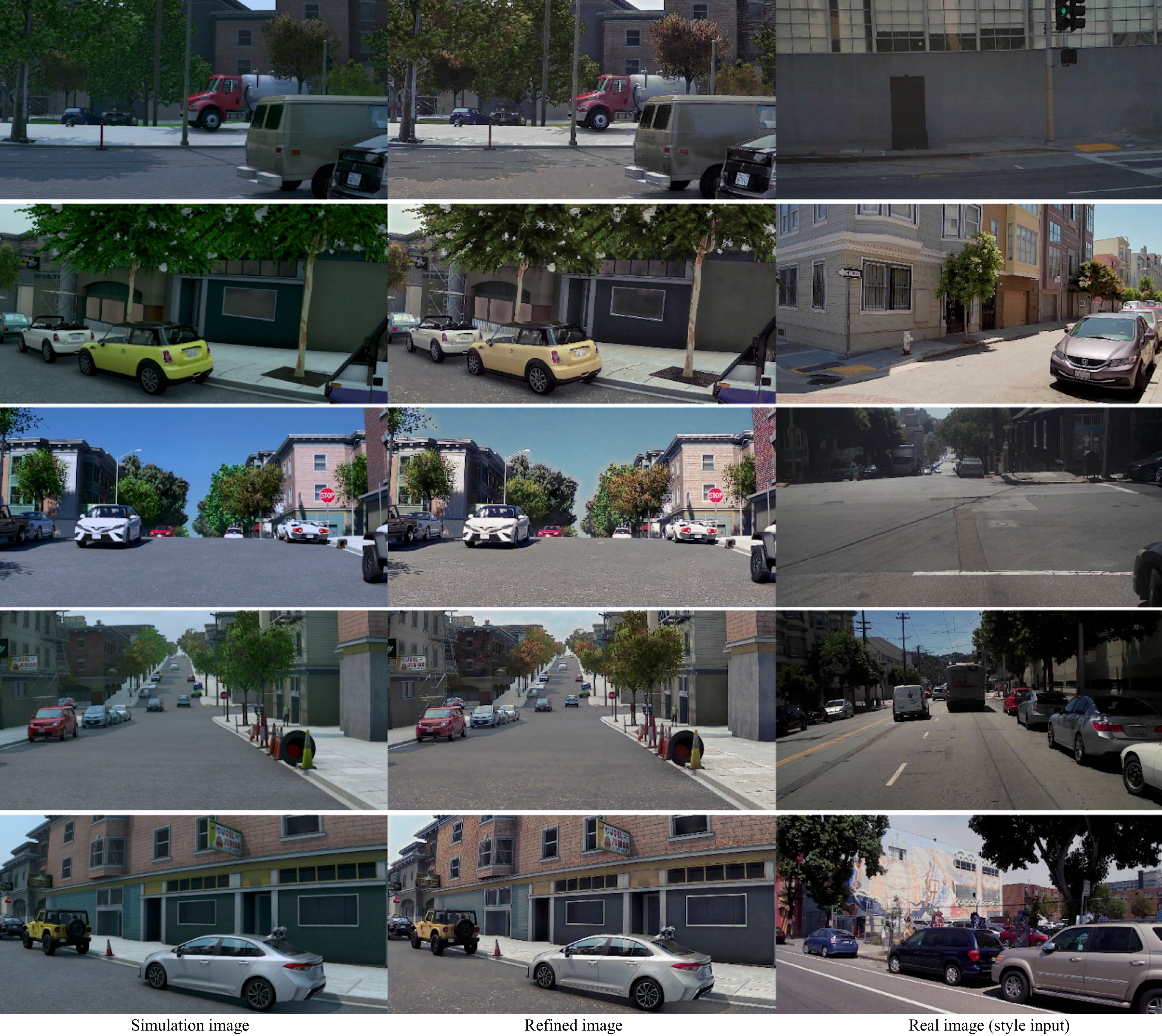}
    \caption{
    \textbf{Qualitative results}. The left column in the figure depicts the simulation images, which exhibit several subtle domain gaps in terms of foliage, road, contrast, sharpness, etc., compared to randomly selected real images in the right column. 
    These real images serve as the unpaired style input during inference.
    Our proposed approach, depicted in the middle column, is capable of effectively bridging the distribution gap and enhancing the realism of the simulated images. The results demonstrate the ability of the AptSim2Real method to bridge the distribution gap and improve the realism of the simulated images. 
    }
    \label{fig:apt_qualitative}
\end{figure*}

\begin{figure*}[htp]
    \centering
    \includegraphics[width=\textwidth]{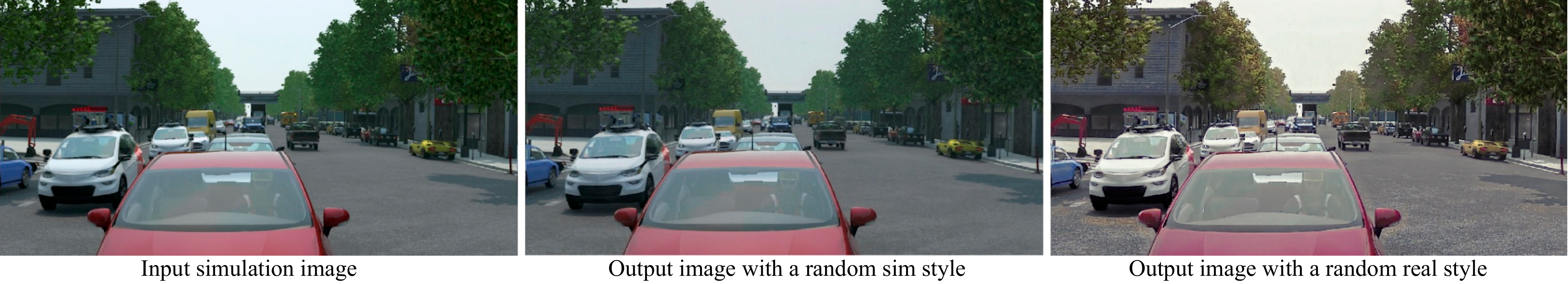}
    \caption{\textbf{Sim style vs real style input}. Examples of generated image using a simulation style image (center) and a real style image (right). By modifying the style code, we are able to control the style of the generated image.}
    \label{fig:sim_vs_real_style}
\end{figure*}

\begin{figure}[htp]
    \includegraphics[width=\textwidth / 2]{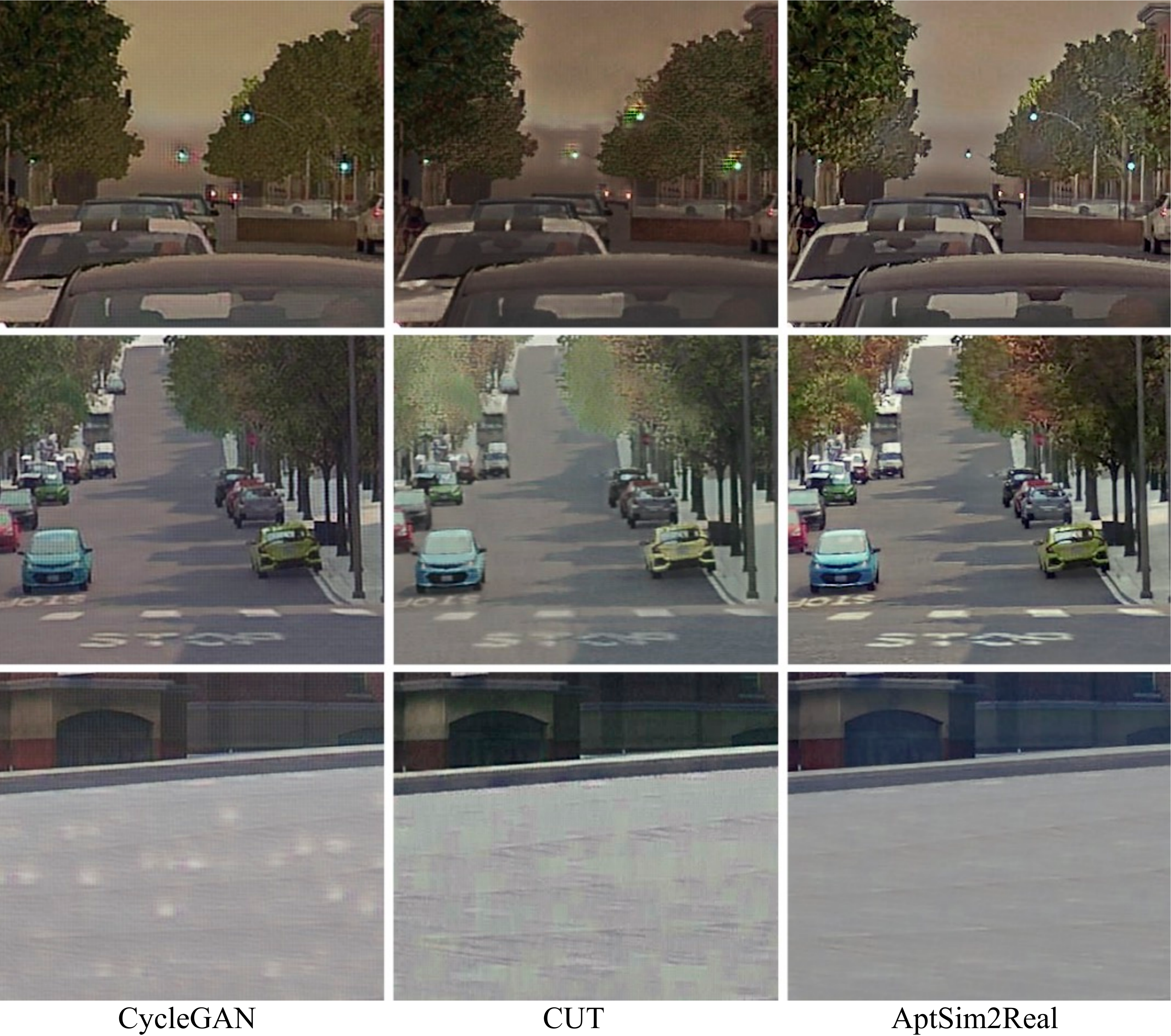}
    \caption{
Comparisons between the proposed AptSim2Real method and other existing approaches such as CycleGAN~\cite{CycleGAN2017} and CUT~\cite{park2020cut} are shown in the figure above. 
We trained the baseline methods with multiple sets of hyper-parameters, and here we are presenting the best results we could obtain.
The results produced by CycleGAN (left column) and CUT (middle column) exhibit multiple artifacts.
For instance, the first row shows high intensity artifacts around the traffic light from both CycleGAN and CUT. 
CycleGAN's generated images also exhibit a noticeable ``checkerboard" pattern artifact, most noticeable in the blue car on the second row. 
CUT results show blurry foliage, as demonstrated in the second row.
Moreover, the road and objects are also blurrier in the baseline methods.
In contrast, our proposed AptSim2Real method, shown in the right column, reduces these artifacts significantly, resulting in much more realistic images. This is also reflected in our quantitative results, which demonstrate the effectiveness of the AptSim2Real method over the other existing approaches.    
\vspace{-0.1in}
    }
    \label{fig:cyclegan_cut_apt}
\end{figure}

\subsection{Experimental Setup}

\textbf{Dataset}. 
The model was trained on a dataset of $\approx 100,000$ real images of outdoor road scenes collected by autonomous vehicles driving in the city of San Francisco. The vehicle contains multiple cameras with a 360-degree view of the scene.
The dataset was labeled with a combination of both automated and human-verified labeling methods, providing rich 3D attributes such as positions of signs, lights, pedestrians, vehicles, and other objects. 
The simulator ingests these real scenes and a procedurally generated 3D map of the city to render high-quality simulation scenes that closely resemble the real-world scenes. 

\textbf{Training setup}. 
Following the standard GAN training process, the generator and discriminator are alternately trained on mini-batches of training samples. 
Unlike other unpaired image translation methods that randomly select samples from both domains, our method prepares batches such that they contain corresponding approximately-paired image pairs.
Our generator is a fully-convolutional network, can be trained on random smaller crops of images (size $256 \times 256$), and can be applied to images of any resolution during inference.

\textbf{Evaluation Setup}. 
Our aim is to evaluate the effectiveness of our method through a test dataset composed of $2000$ pairs of simulated and corresponding approximately-paired images, ($\boldsymbol{x}$, $\boldsymbol{y}$), that were not utilized during training. 
To avoid giving a night-time style image to the daytime simulated input, we limit our selection to those images that closely match the time of day depicted in the simulated input.
The real images, $\boldsymbol{y}$, are only used as a comparison to gauge the generated quality and not as a style input. 
This is because in practical applications, we do not have access to the corresponding real images for each simulated image during inference.
During inference, we randomly sample a real image from the training set ($\boldsymbol{y'}$) to serve as the style input. 
Given a simulated image $\boldsymbol{x}$ from the test dataset and a randomly selected style image $\boldsymbol{y'}$, we generate $\boldsymbol{\hat y} = G(\boldsymbol{x}, S(\boldsymbol{y'}) - S(\boldsymbol{x}))$.
We compare only with the unpaired method, because paired-methods are not applicable due to lack of pixel-wise correspondence.


\subsection{Qualitative Results}




To evaluate the effectiveness of our method, we visually analyzed multiple randomly selected examples as depicted in Figure~\ref{fig:apt_qualitative}. 
The comparison of the images in the middle column with the real images in the third column demonstrates the ability of our approach to accurately capture high-resolution details and variations in noise, texture, and color. 
The proposed method not only improves the overall realism of the scene, but it particularly enhances the realism of foliage, concrete and asphalt textures, as well as the reflections on vehicles. 

Additionally, our method is able to reduce image artifacts common in GAN architectures by leveraging approximately paired data. 
Figure~\ref{fig:cyclegan_cut_apt} shows a comparison of CycleGAN (left), CUT (middle), and APT (right) images. 
We observed that both the CUT and CycleGAN methods suffer from several visual artifacts such as the presence of checkerboard patterns, traffic light distortions, and the ``Tear Drop Artifact"~\cite{Karras2019stylegan2}. 
In contrast, the AptSim2Real method exhibits a significantly reduced occurrence of these artifacts, resulting in a more visually appealing and realistic output.

We also validate that the style input is not ignored by the generator. 
Figure \ref{fig:sim_vs_real_style} displays an example output with a simulation style input and a real style input. 
It is noteworthy that when the input style corresponds to a simulated image, the output appears less realistic in comparison to an input style based on a real image.

\subsection{Quantitative Comparisons to Baselines} \label{quantatative}
To evaluate the performance of our model against previous baseline works, we use Fréchet Inception Distance (FID) scores as a metric. 
FID scores measure the distribution similarity of the intermediate features produced when the real and generated images are processed by the InceptionV3 image network.
A lower FID score indicates that the generated images are more realistic, as the feature map distributions of the generated and real images are more alike. 

Table~\ref{tb:fid_compare} demonstrates the effectiveness of AptSim2Real compared to two state-of-the-art methods: CycleGAN\cite{CycleGAN2017} and CUT\cite{park2020cut}.
As shown in Table~\ref{tb:fid_compare}, the existing unpaired image translation methods are unable to significantly improve the image quality. 
We observe in our qualitative experiments that this is mainly because of artifacts introduced by the models when trained and evaluated on our complex dataset.
We note that the FID score of the baseline methods is comparable to that of the original simulated images. 
However, this similarity can be attributed to the introduction of artifacts by these methods, as demonstrated by our qualitative results in Figure~\ref{fig:cyclegan_cut_apt}.
It is crucial to avoid artifacts when generating synthetic data for downstream model training, because these models can overfit to artifacts and fail to generalize, leading to poor performance in real-world scenarios.
Our model improves the FID score by more than $24\%$ when compared to unpaired methods. 
We also compute the FID score between two different sets of real images to find the lower bound on FID for our datasets.
The FID score between two sets of real images is $6.6$, which means our method is $35\%$ closer to real images than the baseline simulation.

\begin{table}[ht]
\begin{center}
\caption{\label{tb:fid_compare} FID score of the compared methods}
\begin{tabular}{@{}lc@{}}
    \toprule[1.5pt]
    {Method} & {FID score (lower is better)}\\
    \midrule
    Unmodified Sim images & $23.4$ \\
   CycleGAN & $24.7$\\
   CUT & $23.3$\\
   AptSim2Real & $\mathbf{17.6}$\\
    \bottomrule[1.5pt]
\end{tabular}
\end{center}
\end{table}

\subsection{Other Improvements}

While we propose a specific novel architecture that improves on baselines by leveraging approximately paired data, GANs and image-to-image translation are constantly improving through newer architectures and more stable training strategies. 
Approximately-paired data can be leveraged in a variety of ways to improve performance independent of specific network design choices.



\textbf{Projected GANs}. 
To further improve our results, we conducted supplementary experiments utilizing Projected GANs~\cite{Projected_gans}.
In these experiments, the adversarial loss was calculated on image features extracted from the activations of the first encoding layer of a pre-trained object detection model, rather than on the raw RGB image. 
This approach resulted in a noticeable improvement in FID score, which decreased to $16.96$, representing a relative improvement of $4\%$.

\textbf{Multi-Scale Discriminators}. 
In our method, features were extracted from the discriminator at multiple resolutions $(64\times 64, 32\times 32, 16\times 16)$ and the GAN loss was computed at each of these resolutions. 
Unlike traditional approaches that employ separate discriminators for each scale, we sample intermediate feature maps from a single discriminator to increase training efficiency.

\subsection{Additional Training Details}

The model was trained for a maximum of $100,000$ iterations with a batch size of $24$. 
To ensure stability during the training process, one-sided label smoothing was applied to the adversarial loss, as described in~\cite{gantutorial}. 
The ADAM optimizer~\cite{adam} was utilized with beta values of $\beta_1 = 0.5$ and $\beta_2 = 0.99$, and a learning rate of $1e-4$. 
The learning rate remained constant for $20,000$ iterations and then decreased linearly until it reached zero.
Before training the generator and discriminator, the style encoder was pre-trained for one epoch using only the style classification loss with $\lambda^{bce} = 1.0$. 
Afterwards, the generator and the discriminator were then trained $\lambda^{bce} = 0.01$.
For evaluation, we calculated the exponential moving average of the model parameters for both the generator and style encoder, as suggested in ~\cite{gan_ema}. 



\begin{table}[ht]
\begin{center}
\caption{\label{tb:fid_ablation} Ablation study of AptSim2Real}
\begin{tabular}{@{}lc@{}}
    \toprule[1.5pt]
    {Method} & {FID score (lower is better)}\\
    \midrule
    AptSim2Real & $\mathbf{17.6}$ \\
    \midrule
    Removing L1 identity loss & $19.1$ \\
    Removing Luminance loss & $19.2$\\
    AptSim2Real with Unpaired Data & $19.5$\\
    \bottomrule[1.5pt]
\end{tabular}
\end{center}
\end{table} 
\vspace{-0.1in}

\subsection{Ablation Study}

The ablation study in Table~\ref{tb:fid_ablation} show that identity loss helps the model to explicitly capture style information using the style encoder and preventing it from ignoring the style input.
Moreover, we found that the luminance loss effectively reduces the artifacts that are commonly encountered in GAN-based image-to-image networks such as CUT and CycleGAN. 
By enforcing consistency in the average relative luminance between the input and output images, we reduce the bright spot artifacts that would otherwise cause a large difference in average luminance. 
Our experiments also reveal that training AptSim2Real on unpaired data results in a decrease in generated image quality, leading to a relative increase of $10\%$ in the FID score. 
Similarly, when training CUT with approximately paired data, we observed a drop in the FID score, resulting in an FID score of $19.1$. 
These results emphasize the importance of approximately-paired data for achieving high-quality image generation results.

\section{Conclusion}
We introduced a new category of image-to-image translation that can operate with approximately-paired images, making it a practical solution for sim-to-real applications. 
Our proposed architecture takes advantage of this approximately-paired data to produce better results compared to unpaired translation methods.
Our approach integrates the latest advances in GANs and employs multiple loss functions, including a novel luminance loss, to effectively overcome the challenges of sim-to-real transfer and bridge the realism gap. 
\vspace{-0.1in}

\textbf{Acknowledgement:} 
We are grateful to our colleagues Surya Dwarakanath, Luyu Yang, Abhishek Sharma, Ambrish Tyagi, Zhao Chen, and Yuning Chai for their unwavering support and valuable suggestions.


\end{document}